\documentclass[11pt]{article}
\usepackage{amssymb,amsmath, amsthm}
\usepackage{enumerate}
\usepackage{hyperref}
\usepackage{cleveref}
\crefformat{footnote}{#2\footnotemark[#1]#3}
\usepackage[algoruled,boxruled, linesnumbered, shortend, lined]{algorithm2e}
\usepackage{algpseudocode}
\usepackage{cite}
\usepackage{enumitem}
\setenumerate[0]{leftmargin=*}
\usepackage{caption}
\usepackage{graphicx}
\usepackage{float}

\usepackage{mathtools}

\numberwithin{equation}{section}

\newcommand{\inclu}[0] {\ar@{^{(}->}}

\newcommand{\rmi}{\mathrm{i}}

\usepackage[margin=1.1in]{geometry}

\SetAlgoSkip{bigskip}
\SetKwInput{Input}{Input\hspace{34pt}{ }}
\SetKwInput{Output}{Output\hspace{27pt}{ }}
\SetKwInput{Initialize}{Initialize\hspace{2pt}{ }}
\SetKwInput{Parameters}{Parameters\hspace{2pt}{ }}

\newtheorem{theorem}{Theorem}[section]
\newtheorem{proposition}[theorem]{Proposition}
\newtheorem{lemma}[theorem]{Lemma}

\newtheorem{corollary}[theorem]{Corollary}

\title{Training GANs with Centripetal Acceleration}

\begin{document}
	
	\author{Wei Peng\thanks{Department of Mathematics, National University of Defense Technology, Changsha, 410073, Hunan, China.
Email: weipeng0098@126.com}\and Yu-Hong Dai\thanks{Corresponding author.
LSEC, ICMSEC, Academy of Mathematics and Systems Science, Chinese Academy of Sciences, Beijing
100190, China \ \& \ School of Mathematical Sciences, University of Chinese Academy of Sciences, Beijing 100049, China. His research is supported by the 973 Program (No. 2015CB856002) and the Chinese NSF grant (No. 11631013). Email: dyh@lsec.cc.ac.cn}\and Hui Zhang\thanks{Department of Mathematics, National University of Defense Technology, Changsha, 410073, Hunan, China. His research is supported by the Chinese NSF grants (Nos. 11501569 and 61571008). Email: h.zhang1984@163.com}\and Lizhi Cheng
\thanks{Department of Mathematics, National University of Defense Technology, Changsha, 410073, Hunan, China. Email: clzcheng@nudt.edu.cn}
	}	
	
	\date{}	
	\maketitle	
	
	\begin{abstract}
		Training generative adversarial networks (GANs) often suffers from cyclic behaviors of iterates. Based on a simple intuition that the direction of centripetal acceleration of an object moving in uniform circular motion is toward the center of the circle, we present the \textit{Simultaneous Centripetal Acceleration} (SCA) method and the \textit{Alternating Centripetal Acceleration} (ACA) method to alleviate the cyclic behaviors. Under suitable conditions, gradient descent methods with either SCA or ACA are shown to be linearly convergent for bilinear games. Numerical experiments are conducted by applying ACA to existing gradient-based algorithms in a GAN setup scenario, which demonstrate the superiority of ACA.

	\end{abstract}
	
	\bigskip
	
	\textit{Secondary} 90C06, 90C30.
	
	\section{Introduction}
	\textit{Generative Adversarial Nets} (GANs)\cite{goodfellow2014generative} are recognized as powerful generative models, which have successfully been applied to various fields such as image generation\cite{karras2017progressive}, representation learning\cite{radford2015unsupervised} and super resolution\cite{shi2018super}. The idea behind GANs is an adversarial game between a generator network (G-net) and a discriminator network (D-net). The G-net attempts to generate synthetic data from some noise to deceive the D-net while the D-net tries to discern between the synthetic data and the real data. The original GANs can be formulated as the min-max problem:
	\begin{align}\label{prob}
	&\min_{G}\max_{D} V(G,D)=\mathbb{E}_{x\sim p_{data}}[\log D(x)]+\mathbb{E}_{z\sim p_z(z)}[\log(1-D(G(z)))].
	\end{align}
	
	Though GANs are appealing, they are often hard to train. The main difficulty might be the associated gradient vector field rotating around a Nash equilibrium due to the existence of imaginary components in the Jacobian eigenvalues\cite{mescheder2017numerics}, which results in the limit oscillatory behaviors. There are a series of studies focusing on developing fast and stable  methods of training GANs. Using the Jacobian, consensus optimization\cite{mescheder2017numerics} diverts gradient updates to the descent direction of the field magnitudes.
	More essentially, a differential game can always be decomposed into a \textit{potential game} and a \textit{Hamiltonian game}\cite{balduzzi2018mechanics}. Potential games have been intensively studied \cite{monderer1996potential} because gradient decent methods converge in these games. Hamiltonian games obey a conservation law such that iterates generated by gradient descent are likely to cycle or even diverge in these games. Therefore, Hamiltonian components might be the cause of cycling when gradient descent methods are applied. Based on the observations, the \textit{Symplectic Gradient Adjustment} (SGA) method \cite{balduzzi2018mechanics} modifies the associated vector field to guide the iterates to \textit{cross the curl} of the Hamiltonian component of a differential game.
	\cite{gemp2018global} also uses the similar technique to cross the curl such that rotations are alleviated. By augmenting the Follow-the-Regularized-Leader algorithm with an $l_2$ regularizer\cite{shalev2012online} by adding an \textit{optimistic} predictor of the next iteration gradient, \textit{Optimistic Mirror Descent} (OMD) methods are presented in \cite{daskalakis2017training} and analysed in \cite{gidel2018variational,mertikopoulos2018mirror,liang2018interaction,Mokhtari}.
	The negative momentum is employed in \cite{gidel2018negative} to deplete the kinetic energy of the cyclic motion such that iterates would fall towards the center. It is also observed in \cite{gidel2018negative} that the alternating version of the negative momentum method is more stable.

	Our idea is motivated by two aspects. Firstly and intuitively, we use the fact that the direction of centripetal acceleration of an object moving in uniform circular motion points to the center of the circle, which might guide iterates to cross the curl and escape from cycling traps. Secondly, we try to find a method to approximate the dynamics of consensus optimization or SGA to cross the curl without computing the Jacobian, which can reduce computational costs. Then we were inspired to present the centripetal acceleration methods, which can be used to adjust gradients in various methods such as SGD, RMSProp\cite{tieleman2012lecture} and Adam\cite{chilimbi2014project}. For stability and effectiveness, we are also motivated by \cite{gidel2018negative} to study the alternating scheme, which could even work in a notorious GAN setup scenario.
	
	The main contributions are as follows:
	\begin{enumerate}
		\item From two different perspectives, we present centripetal acceleration methods to alleviate the cyclic behaviors in training GANs. Specifically, we propose the \textit{Simultaneous Centripetal Acceleration} (SCA) method and the \textit{Alternating Centripetal Acceleration} (ACA) method.
		\item For bilinear games, which are purely adversarial, we prove that gradient descent with either SCA or ACA is linearly convergent under suitable conditions.
		\item Primary numerical simulations are conducted in a GAN setup scenario, which show that the centripetal acceleration is useful while combining several gradient-based algorithms.
		
	\end{enumerate}

	\textbf{Outline. }The rest of the paper is organized as follows.
	In Section 2, we present simultaneous and alternating centripetal acceleration methods and discuss them with closely related works.
	In Section 3, focusing on bilinear games, we prove the linear convergence of gradient descent combined with the two centripetal acceleration methods.
	In Section 4, we conduct numerical experiments to test the effectiveness of centripetal acceleration methods.
	Section 5 concludes the paper.

	\section{Centripetal Acceleration Methods}
	A differentiable two-player game involves two loss functions $l_1(\theta,\phi)$ and $l_2(\theta,\phi)$ defined over a parameter space $\Omega_\theta\times\Omega_\phi$.
	 Player 1 tries to minimize the loss $l_1$ while player 2 attempts to minimize the loss $l_2$. The goal is to find a local Nash equilibrium of the game, i.e. a pair $ (\bar \theta,\bar \phi)$ with the following two conditions holding in a neighborhood of $(\bar \theta,\bar \phi)$:
	\begin{align*}
	\bar \theta \in\arg\min_{\theta} l_1(\theta,\bar\phi),~~~~\bar \phi\in\arg\min_{\phi} l_2(\bar \theta,\phi).
	\end{align*}
	The derivation of problem \eqref{prob} leads to a two-player game. The G-net is parameterized as $G(\cdot~;\theta)$ while the D-net is parameterized as $D(\cdot~;\phi)$. Then the problem becomes to find a local Nash  equilibrium:
	\begin{align}
	\bar \theta \in\arg\min_{\theta} \left\{V(\theta,\bar\phi)\right\},~~~~\bar \phi\in\arg\min_{\phi} \left\{-V(\bar \theta,\phi)\right\},
	\end{align}
	where
	\begin{align}
	V(\theta,\phi)=	\mathbb{E}_{x\sim p_{data}}[\log D(x;\phi)]+\mathbb{E}_{z\sim p_z(z)}[\log(1-D(G(z;\theta);\phi))].
	\end{align}
	
	The simultaneous gradient descent method in training GANs \cite{nowozin2016f} is
	\begin{align*}
	\theta_{t+1}=\theta_t-\alpha \nabla_{\theta} V(\theta_t,\phi_t),~~~~
	\phi_{t+1}=\phi_t+\alpha \nabla_{\phi} V(\theta_t,\phi_t).
	\end{align*}
	The alternating version is
	\begin{align*}
	\theta_{t+1}=\theta_t-\alpha \nabla_{\theta} V(\theta_t,\phi_t),~~~~
	\phi_{t+1}=\phi_t+\alpha \nabla_{\phi} V(\theta_{t+1},\phi_{t}).
	\end{align*}
	However, directly applying gradient descent even fails to approach the saddle point in a toy model (See Fig. \ref{f1} in Section 4).
	By applying the \textit{Simultaneous Centripetal Acceleration} (SCA) method, which will be explained later, to adjust gradients, we obtain the method of \textit{Gradient descent with SCA}  (Grad-SCA):
	\begin{align}
	G_{\theta} &= \nabla_{\theta} V(\theta_t,\phi_t)+ \frac{\beta_1}{\alpha_1}(\nabla_{\theta}V(\theta_t,\phi_t)-\nabla_{\theta}V(\theta_{t-1},\phi_{t-1}))\label{sim1},\\ \theta_{t+1}&=\theta_t-\alpha_1 G_\theta\label{sim2},\\
	G_{\phi} &= \nabla_\phi V(\theta_t,\phi_t)+ \frac{\beta_2}{\alpha_2}(\nabla_{\phi}V(\theta_t,\phi_t)-\nabla_{\phi}V(\theta_{t-1},\phi_{t-1}))\label{sim3},\\ \phi_{t+1}&=\phi_t+\alpha_2 G_\phi \label{sim4}.
	\end{align}
It can be seen that the gradient decent scheme is still employed in \eqref{sim2} and \eqref{sim4}, while the gradients in \eqref{sim1} and \eqref{sim3} are adjusted by adding the directions of centripetal acceleration simultaneously. If adjusting the gradients by the \textit{Alternating Centripetal Acceleration} (ACA) method,
we obtain the following method of \textit{Gradient descent with ACA} (Grad-ACA):
	\begin{align}
	G_{\theta} &= \nabla_{\theta} V(\theta_t,\phi_t)+ \frac{\beta_1}{\alpha_1}(\nabla_{\theta}V(\theta_t,\phi_t)-\nabla_{\theta}V(\theta_{t-1},\phi_{t-1}))\label{alt1},\\ \theta_{t+1}&=\theta_t-\alpha_1 G_\theta\label{alt2},\\
	G_{\phi} &= \nabla_\phi V(\theta_{t+1},\phi_t)+ \frac{\beta_2}{\alpha_2}(\nabla_{\phi}V(\theta_{t+1},\phi_t)-\nabla_{\phi}V(\theta_{t},\phi_{t-1}))\label{alt3},\\ \phi_{t+1}&=\phi_t+\alpha_2 G_\phi\label{alt4}.
	\end{align}
Grad-ACA also employs simple gradient descent steps but adjusts the gradients by adding the directions of centripetal acceleration alternatively. Nevertheless, the idea of centripetal acceleration can
also be applied to other gradient-based methods, resulting in more
efficient algorithms. For example, the RMSProp algorithm \cite{tieleman2012lecture} with ACA, abbreviated by RMSProp-ACA,
performs well in our numerical experiments (see Section 4.2).

\begin{figure}[H]
		\centering		\includegraphics[width=0.4\textwidth]{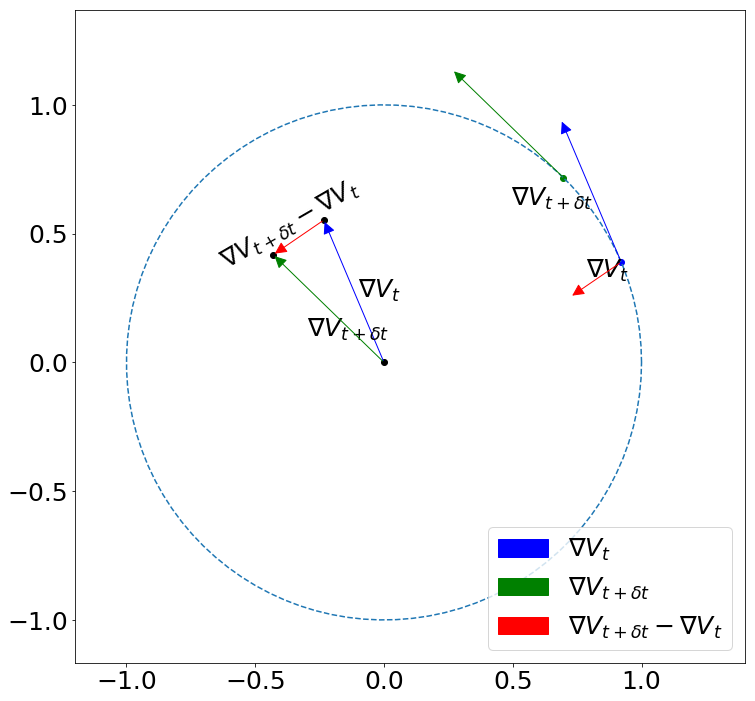}
		\caption{The basic intuition of centripetal acceleration methods.}
		\label{intuition}
	\end{figure}

	The basic intuition behind employing centripetal acceleration is shown in Fig. \ref{intuition}. Consider the uniform circular motion. Let\ $\nabla V_t$ denote the instantaneous velocity at time $t$. Then the centripetal acceleration $\lim_{\delta t\rightarrow 0}(\nabla V_{t+\delta t}-\nabla V_t)/\delta t$ points to the origin. The cyclic behavior around a Nash equilibrium might be similar to the circular motion around the origin. Therefore, the centripetal acceleration provides a direction, along which the iterates can approach the target more quickly. Then the approximated centripetal acceleration term $(\nabla V(\theta_t,\phi_t)-\nabla V(\theta_{t-1},\phi_{t-1}))$ is applied to gradient descent as illustrated in Grad-SCA.	
	The proposed centripetal acceleration methods are also inspired by the dynamics of consensus optimization. In a Hamiltonian game, the associated vector field $\nabla V$ conserves the Hamiltonian's level sets because $\langle\nabla V,\nabla\|\nabla V\|^2\rangle=0$, which prevents iterates from approaching the equilibrium where $\|\nabla V\|=0$. To illustrate the similarity between centripetal acceleration methods and consensus optimization in Hamiltonian games, we consider the $n$-player differential game where each player has a loss function $l_i(w_1,w_2,\cdots,w_n)$ for $i=1,2,\cdots,n$. Then the simultaneous gradient is
	$\xi(w_1,w_2,\cdots,w_n) := (\nabla_{w_1}l_1,\nabla_{w_2}l_2,\cdots,\nabla_{w_n}l_n)$.
	The Jacobian of $\xi$ is
	\begin{align}
	J := \left[
	\begin{array}{cccc}
	\nabla_{w_1w_1} l_1&\nabla_{w_1w_2} l_1&\cdots&\nabla_{w_1w_n} l_1\\
	\nabla_{w_2w_1} l_2&\nabla_{w_2w_2} l_2&\cdots&\nabla_{w_1w_2} l_2\\
	\cdots&\cdots&\cdots&\cdots\\
	\nabla_{w_nw_1} l_n&\nabla_{w_{n}w_2} l_n&\cdots&\nabla_{w_{n}w_n} l_n\\
	\end{array}
	\right].
	\end{align}
	Let $w:=(w_1,w_2,\cdots,w_n)$. Then the iteration scheme of consensus optimization is
	\begin{align}
	w_{k+1}=w_{k}-\alpha(\xi_k+\beta J^T_k\xi_k)
	\end{align}
	and the corresponding continuous dynamics has the form:
	\begin{align}
	\frac{dw}{dt}=-(I+\beta J^T)\xi.
	\end{align}
	When $\beta$ is small, the dynamics approximates
	\begin{align}
	\frac{dw}{dt}=-(I-\beta J^T)^{-1}\xi.
	\end{align}
	By rearranging the order, we obtain
	\begin{align}\label{dy1}
	\frac{dw}{dt}=-\xi+\beta J^T \frac{dw}{dt}.
	\end{align}
	Since the game is assumed to be Hamiltonian, i.e., $J=-J^T$, the dynamic equation \eqref{dy1} becomes
	\begin{align}\label{dy2}
	\frac{dw}{dt}=- \xi-\beta J \frac{dw}{dt}.
	\end{align}
	Note that $J\frac{dw}{dt}=\frac{d\xi}{dt}$. Then \eqref{dy2} is equivalent to
	\begin{align}\label{dy3}
	\frac{dw}{dt}=-\xi-\beta\frac{d\xi}{dt}.
	\end{align}
	Discretizing the equation with stepsize $\alpha$, we obtain
	\begin{align}
	w_{t+1}=w_{t}-\alpha\xi_t-\beta(\xi_t-\xi_{t-1}),
	\end{align}
	which is exactly Grad-SCA. Furthermore, in Hamiltonian games, the dynamics of consensus optimization and SGA that plugs into gradient descent algorithms (Grad-SGA) are essentially the same. Therefore, the presented Grad-SCA could be regarded as a Jacobian-free approximation of consensus optimization or Grad-SGA.

	\textbf{Related works.} Taking $\alpha_1=\alpha_2=\beta_1=\beta_2=\alpha$ in Grad-SCA \eqref{sim1}-\eqref{sim4},  the centripetal acceleration scheme reduces to  OMD\cite{daskalakis2017training}, which has the following form:
	\begin{align*}
	\theta_{t+1}&=\theta_t-2\alpha \nabla_\theta V(\theta_t,\phi_t)+\alpha \nabla_{\theta} V(\theta_{t-1},\phi_{t-1}),\\
	\phi_{t+1}& = \phi_t  +2\alpha \nabla_\phi V(\theta_t,\phi_t)-\alpha \nabla_{\phi} V(\theta_{t-1},\phi_{t-1}).
	\end{align*}
Very recently, from the perspective of generalizing OMD, \cite{Mokhtari} presented schemes similar to Grad-SCA and they studied its convergence under a unified proximal method framework.
	However, OMD is motivated by predicting the next iteration gradient to be the current gradient optimistically. Although the scheme of OMD coincides with Grad-SCA, we must stress that the motivations are essentially different and result in totally distinct parameter selection strategies. Due to the similar dynamics, the presented methods inherit parameter selection strategies of consensus optimization and SGA.  For example, in the second experiment in Section 4, we take $\alpha_1=\alpha_2=5\times 10^{-4}$ and $\beta_1=\beta_2=0.5$. The magnitude of $\beta$ is quite larger than $\alpha$ instead of an equality. Moreover, we analyze the alternating form (Grad-ACA) \eqref{alt1}-\eqref{alt4} and employed RMSProp-ACA in the numerical experiments. Therefore, the presented methods are not trivial generalizations of OMD and the idea of centripetal acceleration is quite useful.
	
	Another similar scheme\cite{gidel2018variational} is to extrapolate the gradient from the past:
	\begin{align*}
	\theta_{t+\frac{1}{2}}&=\theta_t-\alpha\nabla_\theta V(\theta_{t-\frac{1}{2}},\phi_{t-\frac{1}{2}}),&
	\phi_{t+\frac{1}{2}}&=\phi_t+\alpha\nabla_\phi V(\theta_{t-\frac{1}{2}},\phi_{t-\frac{1}{2}}),\\
	\theta_{t+1}&=\theta_t-\alpha\nabla_\theta V(\theta_{t+\frac{1}{2}},\phi_{t+\frac{1}{2}}),&
	\phi_{t+1}&=\phi_t+\alpha\nabla_\phi V(\theta_{t+\frac{1}{2}},\phi_{t+\frac{1}{2}}).
	\end{align*}
	It can be rewritten as
	\begin{align*}
	\theta_{t+\frac{1}{2}}&=\theta_{t-\frac{1}{2}}-2\alpha \nabla_{\theta} V(\theta_{t-\frac{1}{2}},\phi_{t-\frac{1}{2}})+ \alpha\nabla_{\theta}V(\theta_{t-\frac{3}{2}},\phi_{t-\frac{3}{2}}),\\
	\phi_{t+\frac{1}{2}}&=\phi_{t-\frac{1}{2}}+2\alpha \nabla_{\phi} V(\theta_{t-\frac{1}{2}},\phi_{t-\frac{1}{2}})- \alpha\nabla_{\phi}V(\theta_{t-\frac{3}{2}},\phi_{t-\frac{3}{2}})
	\end{align*}
	which is equivalent to OMD. The algorithm may also be closely related to the predictive methods with the following form:
	\begin{align*}
		\theta_{t+\frac{1}{2}}&=\theta_t-\alpha\nabla V(\theta_t,\phi_{t}),&
		\phi_{t+\frac{1}{2}}&=\phi_t+\alpha\nabla V(\theta_t,\phi_{t}),\\
		\theta_{t+1}&=\theta_t-\beta\nabla V(\theta_{t+\frac{1}{2}},\phi_{t+\frac{1}{2}}),&
		\phi_{t+1}&=\phi_t+\beta\nabla V(\theta_{t+\frac{1}{2}},\phi_{t+\frac{1}{2}}).
	\end{align*}
	A unified framework to analyze OMD and predictive methods is presented in \cite{liang2018interaction}.
	
	Last but not least, our idea of using alternating scheme comes from negative momentum methods\cite{gidel2018negative}, which suggests alternating forms might be more stable and effective in practice.
	
	\section{Linear Convergence for Bilinear Games}
	In this section, we focus on the convergence of Grad-SCA and Grad-ACA in the bilinear game:
	\begin{align}
	\min_{\theta\in\mathbb{R}^d}\max_{\phi\in\mathbb{R}^p} \theta^T A\phi+\theta^Tb+c^T\phi,~~~A\in\mathbb{R}^{d\times p},b\in\mathbb{R}^d,c\in\mathbb{R}^p.
	\end{align}
	Any stationary point $(\theta^\ast,\phi^\ast)$ of the game satisfies the first order conditions:
	\begin{align}
	A\phi^\ast+b&=0\\
	A^T\theta^\ast+c&=0.
	\end{align}
		It is obvious that a stationary point exists if and only if $b$ is in the range of $A$ and $c$ is in the range of $A^T$.
	We suppose that such a pair $(\theta^\ast,\phi^\ast)$ exists. Without loss of generality, we shift $(\theta,\phi)$ to $(\theta-\theta^\ast,\phi-\phi^\ast)$. Then the problem is reformulated as:
	\begin{align}
	\min_{\theta\in\mathbb{R}^d}\max_{\phi\in\mathbb{R}^p} \theta^T A\phi,~~~A\in\mathbb{R}^{d\times p}.
	\end{align}
	In the following two subsections, we analyze convergence properties of Grad-SCA and Grad-ACA, respectively. Technique details are postponed to appendices.
	
	\subsection{Linear Convergence of Grad-SCA}
	For the bilinear game, Grad-SCA is specified as
	\begin{align}
	\theta_{t+1}&=\theta_{t}-\alpha_1A\phi_t-\beta_1(A \phi_{t}-A\phi_{t-1})\label{biter1},\\
	\phi_{t+1}&=\phi_{t}+\alpha_2 A^T\theta_t+\beta_2(A^T\theta_{t}-A^T\theta_{t-1})\label{biter2}.
	\end{align}
	Define the matrix $F_1\in\mathbb{R}^{2p+2d}$ as
	\begin{align}
	F_1:=\left[\begin{array}{cccc}
	I_d&-(\alpha_1+\beta_1)A&0&\beta_1A\\
	(\alpha_2+\beta_2)A^T&I_p&-\beta_2A^T&0\\
	I_d&0&0&0\\
	0&I_p&0&0
	\end{array}
	\right].\label{sim-matrix}
	\end{align}
	It is obvious that $[\theta_{t+1},\phi_{t+1},\theta_{t},\phi_{t}]^T=F_1[\theta_{t},\phi_{t},\theta_{t-1},\phi_{t-1}]^T$, where $(\theta_{t},\phi_t)$ are generated by \eqref{biter1} and \eqref{biter2}. For simplicity, we suppose that $A$ is square and nonsingular in Propositions \ref{sim-roots} and \ref{prop1} and Corollary \ref{sim-omd}. Then we prove the linear convergence for a general matrix $A$ in Proposition \ref{sim-gen-con} and Corollary \ref{sim-gen-omd}. We will employ the following well-known lemma to illustrate the linear convergence.
	\begin{lemma}\label{help}
		Suppose that $F\in\mathbb{R}^{p\times p}$ has the spectral radius $\rho(F)<1$. Then the iterative system $x_{k+1}=Fx_{k}$ converges to $0$ linearly. Explicitly, $\forall \varepsilon >0$, there exists a constant $C>0$ such that
		\begin{align}
		\|x_t\|\leq C\left(\rho(F)+\varepsilon\right)^t.
		\end{align}
	\end{lemma}
	\begin{proposition}\label{sim-roots}Suppose that $A$ is square and nonsingular. The eigenvalues of $F_1$ are the roots of the fourth order polynomials:
		\begin{align}
		\lambda^2(1-\lambda)^2+(\lambda(\alpha_2+\beta_2)-\beta_2)(\lambda(\alpha_1+\beta_1)-\beta_1)\zeta,~~~\zeta\in\mathrm{Sp}(A^TA),
		\end{align}
		where $\mathrm{Sp}(\cdot)$ denotes the collection of all eigenvalues.
		
	\end{proposition}
	
	Next, we consider cases when $\alpha_1=\alpha_2=\alpha$ and $\beta_1=\beta_2=\beta$.
	
	\begin{proposition}\label{prop1}
		Suppose that $A$ is square and nonsingular. Then $\Delta_t:=\|\theta_t\|^2+\|\phi_t\|^2+\|\theta_{t+1}\|^2+\|\phi_{t+1}\|^2$ is linearly convergent to 0 if $\alpha$ and $\beta$ satisfy
		\begin{align}
		&0<\alpha+\beta\leq \frac{1}{\sqrt{\lambda_{\max}(A^TA)}},~~~~|\alpha-\beta|\leq \frac{\sqrt{\lambda_{\min}(A^TA)}|\alpha+\beta|^2}{10},
		\end{align}
		where $\lambda_{\max}(\cdot)$ and $\lambda_{\min}(\cdot)$ denote the largest and the smallest eigenvalues, respectively.
	\end{proposition}

	Consider the special case when Grad-SCA reduces to OMD. Then we have the following corollary. The corollary is slightly weaker than the existing result \cite[Lemma 3.1]{liang2018interaction}.
	\begin{corollary}\label{sim-omd}
		Suppose that $A$ is square and nonsingular. If $\alpha_1=\alpha_2=\beta_1=\beta_2=\alpha$ and $0<\alpha\leq\frac{1}{\sqrt{\lambda_{\max}(A^TA)}}$, then $\Delta_t$ is linearly convergent, i.e., $\forall \varepsilon>0$, there exists $C>0$ such that
		\begin{align*}
		{\Delta_{t}}\leq C\left(\varepsilon +\sqrt{\frac{1}{2}+\frac{1}{2}\sqrt{1-\alpha^2\lambda_{\min}(A^TA)}}\right)^{2t}.
		\end{align*}
	\end{corollary}
	Now we do not assume $A$ to be square and nonsingular ($d\geq p$). Instead, suppose $A$ has rank $r$ and the SVD decomposition is $A=UDV^T$, where $D=\mathrm{diag}\{\sigma_1,\sigma_2,\cdots,\sigma_r,0,\cdots,0\}\in\mathbb{R}^{p\times p}$ with $\sigma_1\geq\sigma_2\geq\cdots\geq\sigma_r>0$, $U\in \mathbb{R}^{d\times p}$ and $V\in \mathbb{R}^{p\times p}$. Denote by $M$ the null space of $A$, which means $M=\{x\in\mathbb{R}^p|Ax=0\}$, and by $N$ the null space of $A^T$.
	Note that any $(\tilde \theta,\tilde \phi)\in N\times M$ is a stationary point and we define
	\begin{align*}
	\Delta^P_t := \|\theta_{t+1}-P_N(\theta_{0})\|^2+\|\theta_{t}-P_N(\theta_{0})\|^2+\|\phi_{t+1}-P_M(\phi_{0})\|^2+\|\phi_{t}-P_M(\phi_{0})\|^2,
	\end{align*}
	where $P_N(\cdot)$ denotes the orthogonal projection onto $N$ while $P_M(\cdot)$ denotes the orthogonal projection onto $M$.
	\begin{proposition}\label{sim-gen-con}
		Suppose that $0<\alpha+\beta\leq \frac{1}{\sigma_1}$ and $|\alpha-\beta|/|\alpha+\beta|^2\leq 0.1 \sigma_r$. Then $\Delta^P_t$ is linearly convergent.
	\end{proposition}
	
	With the analogous analysis, we have the following result for OMD.
	\begin{corollary}\label{sim-gen-omd}
		If $\alpha_1=\alpha_2=\beta_1=\beta_2=\alpha$ and $0<\alpha\leq\frac{1}{\sigma_1}$, then $\Delta_t^P$ is linearly convergent, i.e., $\forall \varepsilon>0$, there exists a constant $C>0$ such that
		\begin{align*}
		{\Delta_{t+1}^P}\leq C\left(\varepsilon+\sqrt{\frac{1}{2}+\frac{1}{2}\sqrt{1-\alpha^2\sigma_r^2}}\right)^{2t}.
		\end{align*}
	\end{corollary}
	
	\subsection{Linear Convergence of Grad-ACA}
	In this subsection, we consider Grad-ACA for the bilinear game,
	\begin{align}
	\theta_{t+1}&=\theta_{t}-\alpha_1A\phi_t-\beta_1(A \phi_{t}-A\phi_{t-1})\label{alt-i1},\\
	\phi_{t+1}&=\phi_{t}+\alpha_2 A^T\theta_{t+1}+\beta_2(A^T\theta_{t+1}-A^T\theta_{t})\label{como2}.
	\end{align}
	The update of $\phi_{t+1}$ can be rewritten as:
	\begin{align*}
	\phi_{t+1}&=\phi_t+(\alpha_2+\beta_2)A^T(\theta_{t}-\alpha_1A\phi_t-\beta_1(A \phi_{t}-A\phi_{t-1}))-\beta_2A^T\theta_t.
	\end{align*}
	Thus we define the matrix
	\begin{align}
	F_2:=\left[\begin{array}{cccc}
	I&-(\alpha_1+\beta_1)A&0&\beta_1A\\
	\alpha_2 A^T&I-(\alpha_1+\beta_1)(\alpha_2+\beta_2)A^TA&0&(\alpha_2+\beta_2)\beta_1A^TA\\
	I&0&0&0\\
	0&I&0&0
	\end{array}
	\right],\label{alt-mat}
	\end{align}
	which immediately follows that $[\theta_{t+1},\phi_{t+1},\theta_t,\phi_t]^T=F_2[\theta_t,\phi_t,\theta_{t-1},\phi_{t-1}]^T$.

	\begin{proposition}\label{alt-prop1}
		Suppose that $A$ is square and nonsingular. Consider the special case where $\beta_1=0,\alpha_1=\alpha_2=\beta_2=\alpha$. If $0<\alpha\leq\frac{1}{\sqrt{2\lambda_{\max}(A^TA)}}$, then $\Delta_t:=\|\theta_t\|^2+\|\phi_t\|^2$ is linearly convergent to $0$, i.e., there exists a constant $C>0$ such that
		\begin{align*}
		{\Delta_{t}}\leq C\left(1-\alpha^2\lambda_{\min}(A^TA)+\alpha^4\lambda_{\min}(A^TA)^2\right)^{2t}.
		\end{align*}
	\end{proposition}
	
	Next, we do not assume $A$ to be square and nonsingular. Employing the SVD decomposition $A=UDV^T$ and with the same techniques employed in Proposition \ref{sim-gen-con}, we have
	
	\begin{corollary}
		Consider the special case where $\beta_1=0,\alpha_1=\alpha_2=\beta_2=\alpha$. If $0<\alpha\leq\frac{\sqrt{2}}{2\sigma_1}$, Then $\Delta_t^P:=\|\theta_t-P_N(\theta_0)\|^2+\|\phi_t-P_M(\phi_0)\|^2$ is linearly convergent, i.e., there exists a constant $C>0$ such that
		\begin{align*}
		{\Delta_{t}^P}\leq C(1-\alpha^2\sigma_r^2+\alpha^4\sigma_r^4)^{2t},
		\end{align*}
		which implies that $(\theta_t,\phi_t)$ linearly converges to the stationary point $(P_N(\theta_0),P_M(\phi_0))$.
	\end{corollary}

	\section{Numerical Simulation}
	
	\subsection{A Simple Bilinear Game}
\begin{figure}[H]
		\centering
		\includegraphics[width=0.4\textwidth]{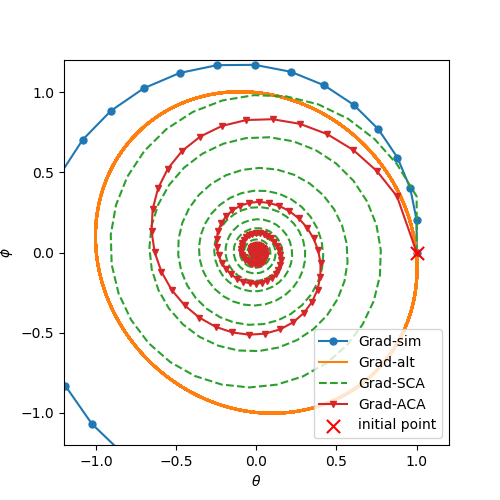}
		\caption{The effects of Grad-SCA and Grad-ACA in the simple bilinear game. Simultaneous gradient descent ($\alpha=0.1,\beta=0$) diverges while the alternating gradient descent ($\alpha=0.1,\beta=0$) keeps the iterates running on a closed trajectory. Instead, both Grad-SCA and Grad-ACA ($\alpha=0.1,\beta=0.3$) converge to the origin linearly and the alternating version seems faster.}
		\label{f1}
	\end{figure}

	In the first experiment, we tested Grad-SCA and Grad-ACA on the following bilinear game
	\begin{align}
	\min_{\theta\in\mathbb{R}}\max_{\phi\in\mathbb{R}} \theta\cdot\phi.
	\end{align}
	The unique stationary point is $(\theta^\ast,\phi^\ast)=(0,0)$. The behaviors of the methods are presented in Fig. \ref{f1}. Pure gradient descent steps do not converge to the origin in this simple game. However, with centripetal acceleration methods, both Grad-SCA and Grad-ACA converge to the origin.
	
\begin{figure}[H]
		\centering
		\includegraphics[width=0.7\textwidth]{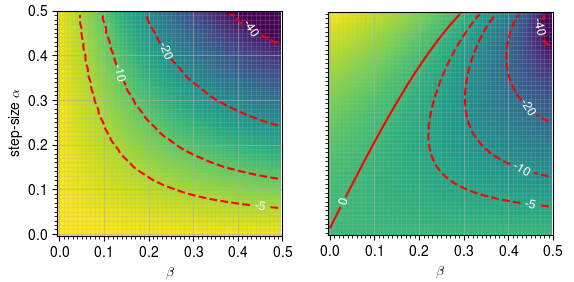}
		\caption{Parameter selection in the simple bilinear game. We test Grad-SCA and Grad-ACA with varying parameters $(\alpha,\beta)\in(0,0.5]\times(0,0.5]$. Each grid point represents the logarithm of the squared distance to the origin after $500$ iterations. Note that the colormaps are different between the two images. Grad-ACA (left) converges in the entire parameter box while Grad-SCA (right) might diverge if the step-size $\alpha$ is much larger than $\beta$. In this simple experiment, a larger $\beta$ seems more preferred. Particularly, when $\alpha=\beta$, the Grad-SCA reduces to OMD.}
		\label{f2}
	\end{figure}

	We compared the effects of various step-sizes and acceleration coefficients in both simultaneous and alternating cases. Fig. \ref{f2} suggests that the alternating methods are preferable.
	\subsection{Mixture of Gaussians}	
\begin{figure}[H]
		\centering
		\includegraphics[width=0.3\textwidth]{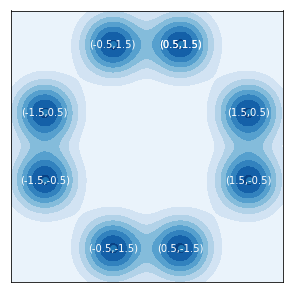}
		\caption{Kernel density estimation on 2560 samples of the ground truth.}
		\label{f3}
	\end{figure}

In the second simulation\footnote{The code is available at \url{https://github.com/dynames0098/GANsTrainingWithCenAcc}}, we established a toy GAN model to compare several methods on learning eight Gaussians with standard deviation $0.04$. The ground truth is shown in Fig. \ref{f3}.

	Both the generator and the discriminator networks have four fully connected layers of $256$ neurons. Each of the four layers is activated by a ReLU layer. The generator has two output neurons to represent a generated point while the discriminator has one output which judges a sample. The random noise input for the generator is a 16-D Gaussian. We conducted the experiment on a server equipped with CPU i7 4790, GPU Titan Xp, 16GB RAM as well as TensorFlow (version 1.12) and Python (version 3.6.7).

\begin{figure}[H]
		\centering
		\includegraphics[width=1\textwidth]{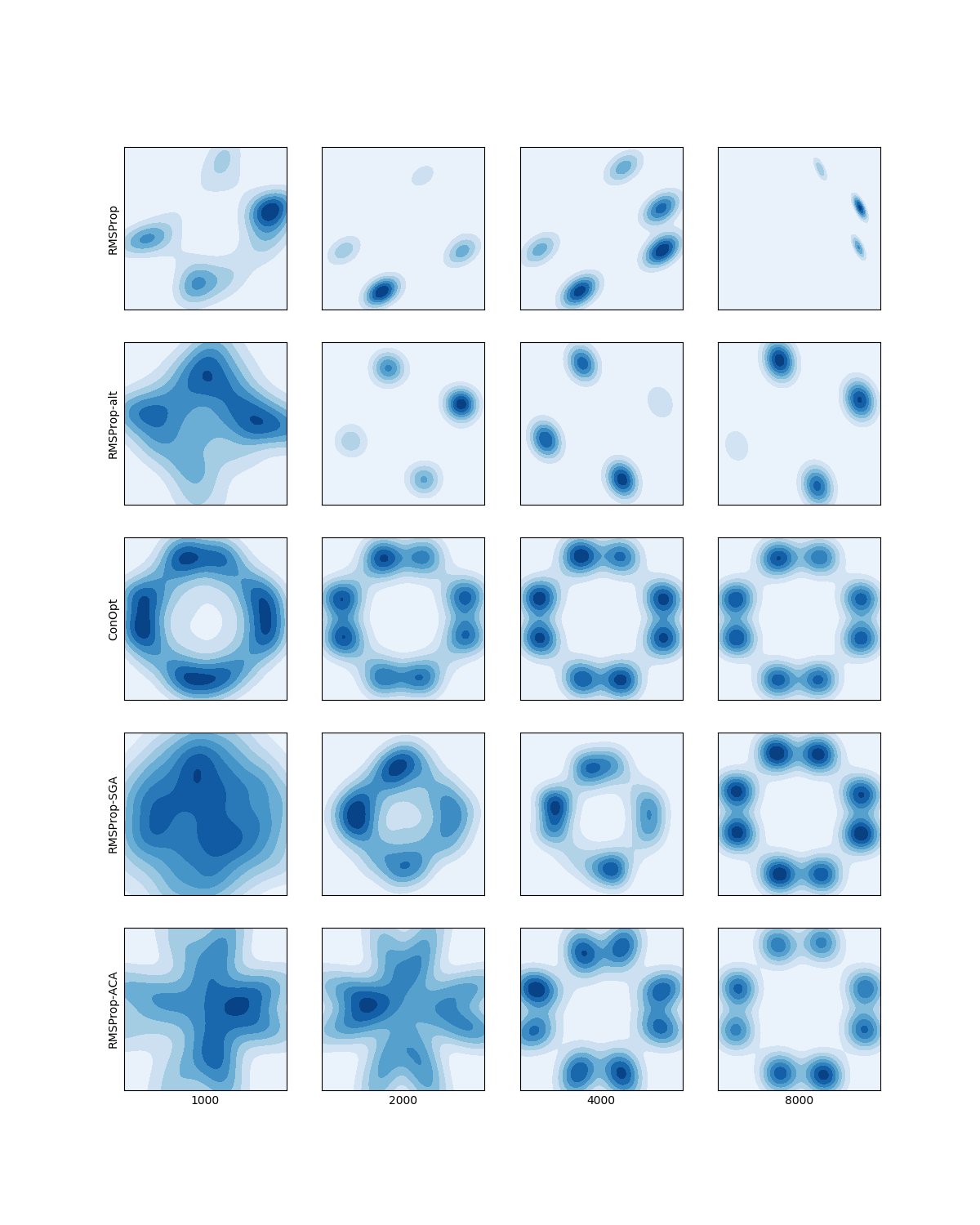}
		\caption{Comparison among several algorithms on the mixture of Gaussians. Five methods are included in the comparison. Each row displays one method and each column shows samples generated by the G-net at $1000,2000,4000,8000$ iterations respectively.}
		\label{f4}
	\end{figure}

	We compared the results of several algorithms as shown in Fig. \ref{f4}. Five methods are included in the comparison:
	\begin{enumerate}
		\item \textbf{RMSProp}: Simultaneous RMSPropOptimizer (learning rate: $\alpha=5\times 10^{-4}$) provided by TensorFlow.
		\item \textbf{RMSProp-alt}: Alternating RMSPropOptimizer (learning rate:  $\alpha=5\times 10^{-4}$).
		\item \textbf{ConOpt}: Consensus optimizer ($h=10^{-4},\gamma=1$)\cite{mescheder2017numerics}.
		\item \textbf{RMSProp-SGA}: Symplectic gradient adjusted RMSPropOptimizer with sign alignment ($\text{learning rate}=10^{-4},\xi=1$)\cite{balduzzi2018mechanics}.
		\item \textbf{RMSProp-ACA}: RMSPropOptimizer with alternating centripetal acceleration method ($\alpha=5\times10^{-4},\beta=0.5$).
	\end{enumerate}
	To stress the effectiveness brought by parameter selection and alternating strategy regardless of the similar form with OMD, we also tested OMD on this simulation with searching a range of parameters (See Appendix \ref{OMDtest}).
	
\begin{figure}[H]
		\centering
		\includegraphics[width=0.6\textwidth]{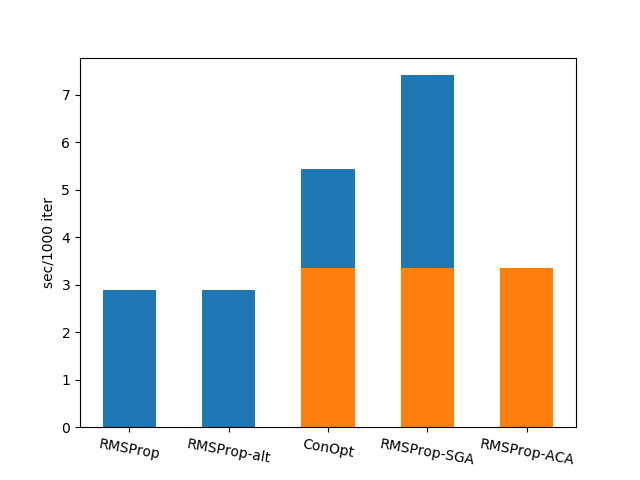}
		\caption{Time consuming comparison. RMSProp-ACA consumes slightly more time than RMSProp. However, it takes far less time than ConOpt and RMSProp-SGA.}
		\label{f5}
	\end{figure}

	The centripetal acceleration methods have extra computation costs on computing the difference between successive gradients as well as storage costs to maintain previous gradients. The consensus optimization and SGA require extra computations on the Jacobian related steps. Fig. \ref{f5} shows a time consuming comparison.
	From these comparisons, RMSProp-ACA seems competitive to other methods.
		
	\section{Conclusion}
	In this paper, to alleviate the difficulty in finding a local Nash equilibrium in a smooth two-player game, we were inspired to present several gradient-based methods, including Grad-SCA and Grad-ACA, which employ centripetal acceleration. The proposed methods can easily be plugged into other gradient-based algorithms like SGD, Adam or RMSProp in both simultaneous or alternating ways. From the theoretical viewpoint, we proved that both Grad-SCA and Grad-ACA have linear convergence for bilinear games under suitable conditions. We found that in a simple bilinear game, centripetal acceleration makes iterates converge to the Nash equilibrium stably; these examples also suggest that alternating methods are more preferred than simultaneous ones. In the GAN setup numerical simulations, we showed that the RMSProp-ACA can be competitive to consensus optimization and symplectic gradient adjustment methods.
	
	However, we only consider the deterministic bilinear games theoretically and limited numerical simulations. In practical training of GANs or its variants, the associated games are much more complicated due to the randomness of computation, the online procedure and non-convexity. These issues still need further detailed studies.
	
	\bibliographystyle{plain}
	\bibliography{CenRelated}
	\appendix

	\section{Proofs in Section 3}
	\subsection{Proof of Proposition \ref{sim-roots}}
	\textbf{Proof} The characteristic polynomial of the matrix \eqref{sim-matrix} is
	\begin{align}\det
	\left(\begin{array}{cccc}
	I_d-\lambda I_d&-(\alpha_1+\beta_1)A&0&\beta_1A\\
	(\alpha_2+\beta_2)A^T&I_p-\lambda I_p&-\beta_2A^T&0\\
	I_d&0&-\lambda I_d&0\\
	0&I_p&0&-\lambda I_p
	\end{array}
	\right ),
	\end{align}
	which is equivalent to
	\begin{align}
	\det \left(\begin{array}{cc}
	\lambda(1-\lambda)I_d&\lambda(\alpha_1+\beta_1)A-\beta_1A\\
	-\lambda(\alpha_2+\beta_2)A^T+\beta_2A^T&\lambda(1-\lambda)I_p
	\end{array}\right)\label{sim-22}.
	\end{align}
	Since $A$ is nonsingular and square, then $0$ or $1$ can not be the roots of \ref{sim-22}. Then the roots of \eqref{sim-22} must be the roots of
	\begin{align}
	\det\left(\lambda(1-\lambda)I_p+\frac{1}{\lambda(1-\lambda)}\left(\lambda(\alpha_2+\beta_2)-\beta_2\right)(\lambda(\alpha_1+\beta_1)-\beta_1)A^TA\right).
	\end{align}
	It follows that the eigenvalues of $F_1$ must be the roots of the fourth order polynomials:
	\begin{align*}
	\lambda^2(1-\lambda)^2+(\lambda(\alpha_2+\beta_2)-\beta_2)(\lambda(\alpha_1+\beta_1)-\beta_1)\zeta,~~~\zeta\in\mathrm{Sp}(A^TA).
	\end{align*}
	\qed
	
	\subsection{Proof of Proposition \ref{prop1}}
	\textbf{Proof} Given an eigenvalue $\lambda$ of $F_1$, using Proposition \ref{sim-roots}, we have
	\begin{align}\label{simueq}
	\left(\lambda^2-\lambda-\mathrm{i}(\lambda(\alpha+\beta)-\beta)\sqrt{\zeta}\right)\left(\lambda^2-\lambda+\mathrm{i}(\lambda(\alpha+\beta)-\beta)\sqrt{\zeta}\right)=0,~~~\zeta\in\mathrm{Sp}(A^TA).
	\end{align}
	Denote $s:=\alpha\sqrt\zeta+\beta\sqrt\zeta$ and $t:=\alpha\sqrt{\zeta}-\beta\sqrt\zeta$. Then the four roots of
	\eqref{simueq} are
	\begin{align*}
	\lambda^{\pm}_1 & = \frac{1+\rmi s\pm\sqrt{1+2\rmi t-s^2}}{2},\\
	\lambda^{\pm}_2 & = \frac{1-\rmi s\pm\sqrt{1-2\rmi t-s^2}}{2}.
	\end{align*}
	Note that for a given complex number $z$, the absolute value of the real part of $z^{\frac{1}{2}}$ is $\sqrt{\frac{|z|+R(z)}{2}}$ and the absolute value of the imaginary part of $z^{\frac{1}{2}}$ is $\sqrt{\frac{|z|-R(z)}{2}}$. Therefore, since $s\leq1$,  all real parts of $\lambda_{i}^\pm(i=1,2)$ lie in the interval $[-\mathcal{R},\mathcal{R}]$, where
	\begin{align}
	\mathcal{R}=\frac{1}{2}\sqrt \frac{\sqrt{(1-s^2)^2+4t^2}+1-s^2}{2}+\frac{1}{2}\label{R}
	\end{align}
	and all imaginary parts of $\lambda_{i}^\pm(i=1,2)$ lie in the interval $[-\mathcal{I},\mathcal{I}]$, where
	\begin{align}
	\mathcal{I}=\frac{1}{2}\sqrt \frac{\sqrt{(1-s^2)^2+4t^2}-1+s^2}{2}+\frac{s}{2}.\label{I}
	\end{align}
	Using the inequality
	\begin{align}
	\sqrt{x+y}\leq \sqrt{x}+\frac{y}{2\sqrt x},~~~~(x>0,y\geq0),\label{ineq}
	\end{align}
	we have
	\begin{align}
	\mathcal{R}&\leq \frac{1}{2}\sqrt{1-s^2+\frac{t^2}{1-s^2}}+\frac{1}{2},\label{r1}\\
	\mathcal{I}&\leq \frac{s}{2}+\frac{|t|}{2\sqrt{1-s^2}}\label{i1}.
	\end{align}
	Next, we discuss $s$ in $(0,1/\sqrt{2}]$ and $(1/\sqrt{2},1]$ separately.\\
	(1). In the first case, we suppose $0<s\leq 1/\sqrt{2}$. Since $|\alpha-\beta|/(\alpha+\beta)^2\leq 0.1\sqrt{\zeta}$ for all $\zeta\in \mathrm{Sp}(A^TA)$, we have
	\begin{align*}
	|t|&\leq \frac{s^2}{10}.
	\end{align*}
	Noting that $\frac{s^2}{2}\leq 1-\sqrt{1-s^2}$, we obtain
	\begin{align}
	|t|\leq \frac{1-\sqrt{1-s^2}}{5}\label{r2}.
	\end{align}
	Combining $s\leq 1/\sqrt{2}$ and \eqref{r2} yields
	\begin{align*}
	|t|&\leq \frac{2(1-\sqrt{1-s^2})(1-s^2)}{5}\\
	&\leq \frac{(1-\sqrt{1-s^2})(1-s^2)}{2\sqrt{1-s^2}+\frac{1}{2}},
	\end{align*}
	which follows that
	\begin{align}
	1&\geq\frac{|t|}{\sqrt{1-s^2}}+\sqrt{1-s^2}+\frac{|t|}{2(1-s^2)}+\frac{|t|}{\sqrt{1-s^2}}\nonumber\\
	&\geq \frac{t^2}{1-s^2}+\sqrt{1-s^2}+\frac{t^2}{2(1-s^2)^{\frac{3}{2}}}+\frac{s|t|}{\sqrt{1-s^2}}\label{what1}\\
	&\geq \frac{t^2}{1-s^2}+\sqrt{1-s^2+\frac{t^2}{1-s^2}}+\frac{s|t|}{\sqrt{1-s^2}}\label{what2}.
	\end{align}
	The inequality \eqref{what1} follows by the fact that $|t|/\sqrt{1-s^2}\leq s/\sqrt{1-s^2}\leq1$ and the inequality \eqref{what2} uses \eqref{ineq}. The inequality above is equivalent to
	\begin{align*}
	\left(\frac{1}{2}\sqrt{1-s^2+\frac{t^2}{1-s^2}}+\frac{1}{2}\right)^2+\left(\frac{s}{2}+\frac{|t|}{2\sqrt{1-s^2}}\right)^2\leq 1.
	\end{align*}
	Using \eqref{r1} and \eqref{i1}, we obtain
	\begin{align}
	\rho(F_1)\leq \sqrt{\mathcal{R}^2+\mathcal{I}^2}\leq 1.\label{sim-final}
	\end{align}
	Note that the equality of \eqref{ineq} holds if and only if $y=0$. Thus the equality of \eqref{sim-final} implies $t=0$ and $s=0$. Since $s>0$, we have the strict inequality $\rho(F_1)<1$, which leads to the linear convergence of $\Delta_t$.\\
	(2). In the second case, assume $1/\sqrt{2}< s\leq1$. Since $t\leq s^2/10\leq0.1$, using \eqref{R} and \eqref{I} directly, we have
	\begin{align}
	\rho(F_1)\leq\sqrt{\mathcal{R}^2+\mathcal{I}^2}<1,
	\end{align}
	which yields the linear convergence.
	\qed
	
	\subsection{Proof of Corollary \ref{sim-omd}}
	\textbf{Proof}
	For the special cases, we have $t=0$ and $0<s\leq 1$.
	From \eqref{r1} and \eqref{i1}, we obtain
	\begin{align}
	\rho(F_1)\leq\sqrt{\mathcal{R}^2+\mathcal{I}^2}&=\frac{1}{2}\sqrt{\left((1-s^2)+1+2\sqrt{1-s^2}\right)+s^2}\nonumber\\
	&=\sqrt{\frac{1}{2}+\frac{1}{2}\sqrt{1-s^2}}\nonumber\\
	&\leq \sqrt{\frac{1}{2}+\frac{1}{2}\sqrt{1-\alpha^2\lambda_{\min}(A^TA)}  } <1.\nonumber
	\end{align}
	From Lemma \ref{help} it follows that $\Delta_t$ is linearly convergent.
	\qed
	
	\subsection{Proof of Proposition \ref{sim-gen-con}}
	\textbf{Proof}
	Using the SVD decomposition $A=UDV^T$, we have
	\begin{align*}
	U^T\theta_{t+1}&=U^T\theta_t-\alpha DV ^T\phi_{t} -\beta (DV^T\phi_{t}-DV^T\phi_{t-1}),\\
	V^T\phi_{t+1}&=V^T\phi_t+\alpha DU^T \theta_{t} +\beta (DU^T\theta_{t}-DU^T\theta_{t-1}) .
	\end{align*}
	According to the definition of the diagonal matrix $D$, the $(r+1)$-th component to $p$-th components of $DU^T\theta_t$ and $DV^T\phi_t$ are zeros. Therefore, we focus on the leading $r$ components of $U^T\theta_t$ and $V^T\phi_t$, denoted by $[U^T\theta_t]_{1:r}$ and $[V^T\phi_t]_{1:r}$ respectively. Let $D_r$ be the matrix composed of the leading $r$ rows and columns of $D$. Then we have
	\begin{align}
	[U^T\theta_{t+1}]_{1:r}&=[U^T\theta_t]_{1:r}-\alpha D_r[V ^T\phi_{t}]_{1:r} -\beta (D_r[V^T\phi_{t}]_{1:r}-D_r[V^T\phi_{t-1}]_{1:r}),\label{comp1}\\
	[V^T\phi_{t+1}]_{1:r}&=[V^T\phi_t]_{1:r}+\alpha D_r[U^T \theta_{t}]_{1:r} +\beta (D_r[U^T\theta_{t}]_{1:r}-D_r[U^T\theta_{t-1}]_{1:r}).\label{comp3}
	\end{align}
	Define
	\begin{align}
	\Delta^r_t:&=\|[U^T\theta_t]_{1:r}\|^2+\|[U^T\theta_{t+1}]_{1:r}\|^2+\|[V^T\phi_t]_{1:r}\|^2+\|[V^T\phi_{t+1}]_{1:r}\|^2\nonumber\\
	&=\|\theta_t- P_N(\theta_t)\|^2+\|\theta_{t+1}- P_N(\theta_{t+1})\|^2+\|\phi_t-P_M(\phi_t)\|^2+\|\phi_{t+1}- P_M(\phi_{t+1})\|^2.\label{eq}
	\end{align}
	The equality \eqref{eq} holds due to $N^{\perp}=\mathrm{Span}\{u_1,u_2,\cdots,u_r\}$ and $M^{\perp}=\mathrm{Span}\{v_1,v_2,\cdots,v_r\}$.
	Since $D_r$ is square and nonsingular, applying Proposition \ref{prop1} to \eqref{comp1} and \eqref{comp3}, we have that $\Delta_t^r$ is linearly convergent.
	Recalling \eqref{biter1}, \eqref{biter2}, for all $u\in N,v\in M$, we have
	\begin{align*}
	u^T\theta_{t+1}&=u^T\theta_t,\\
	v^T\phi_{t+1}&=v^T\phi_t,
	\end{align*}
	which implies $P_N(\theta_t)=P_N(\theta_0)$ and $P_M(\phi_t)=P_M(\phi_0)$ for all $t\geq0$. Then we have $\Delta_t^r=\Delta_t^P$ for all $t\geq 0$. Thus $\Delta_t^P$ is also linearly convergent.\qed
	
	\subsection{Proof of Proposition \ref{alt-prop1}}
	\textbf{Proof} The iterations \eqref{alt-i1} and \eqref{como2} are simplified to
	\begin{align*}
	\theta_{t+1} & =\theta_t-\alpha A\phi_t,\\
	\phi_{t+1} & =(I-2\alpha^2 A^TA)\phi_t+\alpha A^T\theta_t.
	\end{align*}
	Then the matrix $F_2$ reduces to
	\begin{align*}
	\tilde F_2=\left[\begin{array}{cc}
	I&-\alpha A\\
	I-2\alpha^2A^TA&\alpha A^T
	\end{array}\right]
	\end{align*}
	with $[\theta_{t+1},\phi_{t+1}]^T=\tilde F_2[\theta_t,\phi_t]^T$ holding. Eigenvalues of $\tilde F_2$ satisfy
	\begin{align*}
	\lambda^2-(1-2\alpha^2\zeta)\lambda-\alpha^2\zeta=0,~~\zeta\in \mathrm{Sp}(A^TA).
	\end{align*}
	Let $a:=\alpha\sqrt{\zeta} $. Then the two roots are
	\begin{align}
	\lambda=\frac{1-2a^2\pm \sqrt{1+4a^4}}{2}.\label{alt-roots}
	\end{align}
	For all $a\in(0, 1/\sqrt{2})$, applying \eqref{ineq} to \eqref{alt-roots}  we have
	\begin{align*}
	|\lambda|< 1-a^2+a^4.
	\end{align*}
	Note that $f(x)=1-x^2+x^4$ is monotone decreasing on $(0,1/\sqrt{2})$. Then
	\begin{align*}
	\rho(\tilde F_2)< 1-\alpha^2\lambda_{\min}(A^TA)+\alpha^4\lambda_{\min}(A^TA)^2<1.
	\end{align*}
	Therefore, $\Delta_t$ is linearly convergent.\qed	
	\section{Performance of OMD on Mixture of Gaussians}\label{OMDtest}
\begin{figure}[H]
		\centering
		\includegraphics[width=0.8\textwidth]{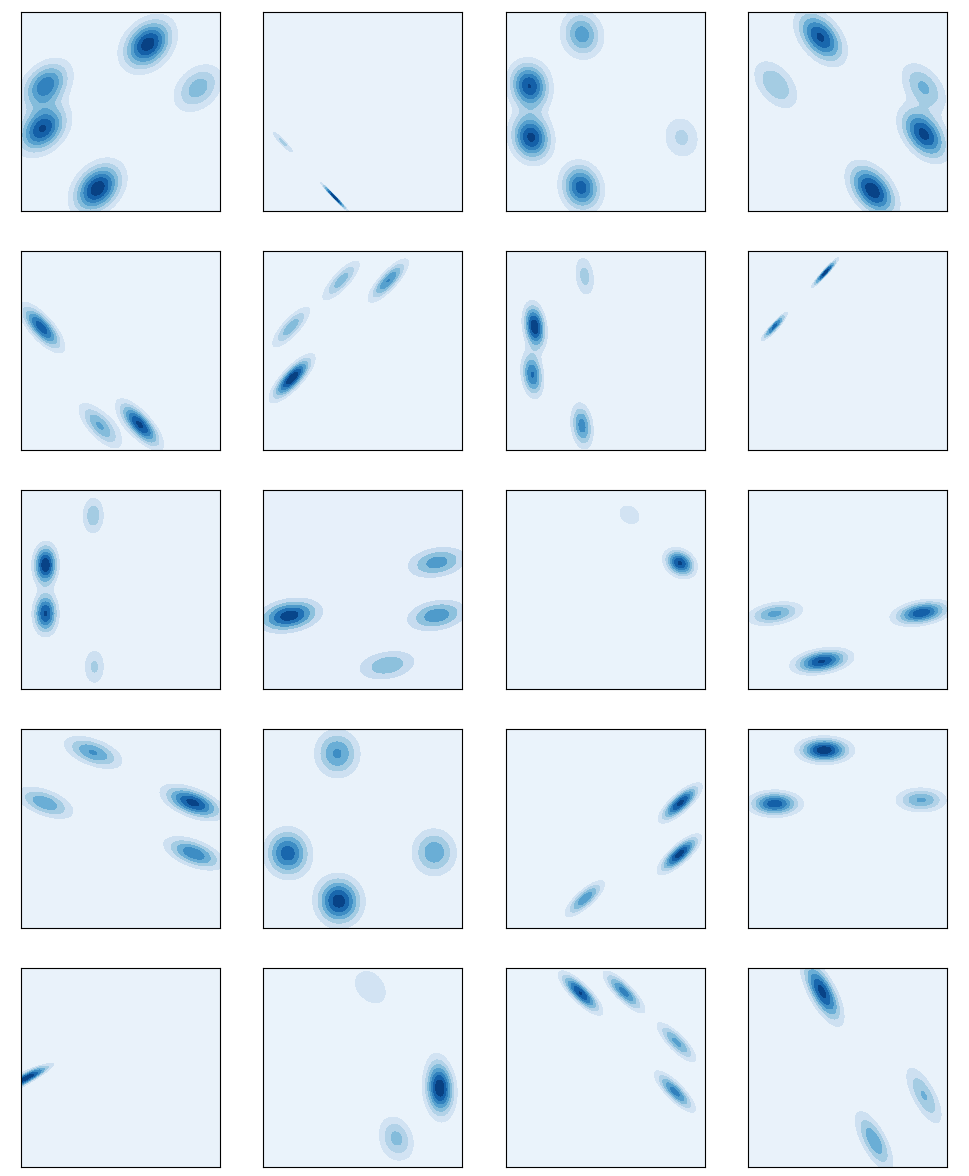}
		\caption{Performance of OMD. From left to right, top to bottom, the stepsize $\alpha$ successively takes values from [2E-5, 5E-5, 1E-4, 2E-4, 5E-4, 1E-3, 2E-3, 5E-3, 1E-2, 2E-2, 5E-2, 0.1, 0.2, 0.5, 1, 2, 5, 10, 20, 50].}
		\label{omd}
	\end{figure}

	For the performance of OMD in the second experiment in Section 4, we search the learning rates on the grid from $0.00002$ to $50$. The result is as Fig. \ref{omd} shows. With varying learning rates, OMD combined with RMSProp suffers from mode collapse and fails to recover the Gaussian mixture even after 20k iterations.

\end{document}